%% file: ms.tex
\documentclass[10pt,twocolumn,letterpaper]{article}

\usepackage{iccv}
\usepackage{times}
\usepackage{epsfig}
\usepackage{graphicx}
\usepackage{amsmath}
\usepackage{amssymb}

\usepackage{units}
\usepackage{subcaption}
\usepackage{multirow}
\usepackage{bm}
\usepackage{mydefs}
\usepackage{bbm}
\usepackage{pifont}

\usepackage[pagebackref=true,breaklinks=true,letterpaper=true,colorlinks,bookmarks=false]{hyperref}

\iccvfinalcopy 

\def\iccvPaperID{5412} 

\ificcvfinal\fi
\begin{document}

\newcommand{\supsecref}[1]{\secref{#1}}
\newcommand{\supfigref}[1]{\figref{#1}}
\newcommand{\suptblref}[1]{\tblref{#1}}

\title{Patchwork: A Patch-wise Attention Network for \\ Efficient Object Detection and
Segmentation in Video Streams}


\author{Yuning Chai\thanks{Now at Waymo LLC.}\\
Google Inc.\\
{\tt\small chaiy@google.com}
}

\maketitle

\input{main_content}

\PAR{Acknowledgments.} We like to thank the anonymous reviewers, Henrik Kretzschmar, Benjamin Sapp and Scott Ettinger for their helpful comments.

{\small
\bibliographystyle{ieee}
\bibliography{refs}
}

\clearpage

\input{supp_content}

\end{document}

%% file: main_content.tex

\begin{abstract}

Recent advances in single-frame object detection and segmentation techniques have motivated a wide range of works to extend these methods to process video streams. In this paper, we explore the idea of hard attention aimed for latency-sensitive applications. Instead of reasoning about every frame separately, our method selects and only processes a small sub-window of the frame. Our technique then makes predictions for the full frame based on the sub-windows from previous frames and the update from the current sub-window. The latency reduction by this hard attention mechanism comes at the cost of degraded accuracy. We made two contributions to address this. First, we propose a specialized memory cell that recovers lost context when processing sub-windows. Secondly, we adopt a Q-learning-based policy training strategy that enables our approach to intelligently select the sub-windows such that the staleness in the memory hurts the performance the least. Our experiments suggest that our approach reduces the latency by approximately four times without significantly sacrificing the accuracy on the ImageNet VID video object detection dataset and the DAVIS video object segmentation dataset. We further demonstrate that we can reinvest the saved computation into other parts of the network, and thus resulting in an accuracy increase at a comparable computational cost as the original system and beating other recently proposed state-of-the-art methods in the low latency range.

\end{abstract}

\section{Introduction}

\begin{figure*}[!tbp]
  \begin{subfigure}[b]{0.72\textwidth}
    \includegraphics[width=\textwidth]{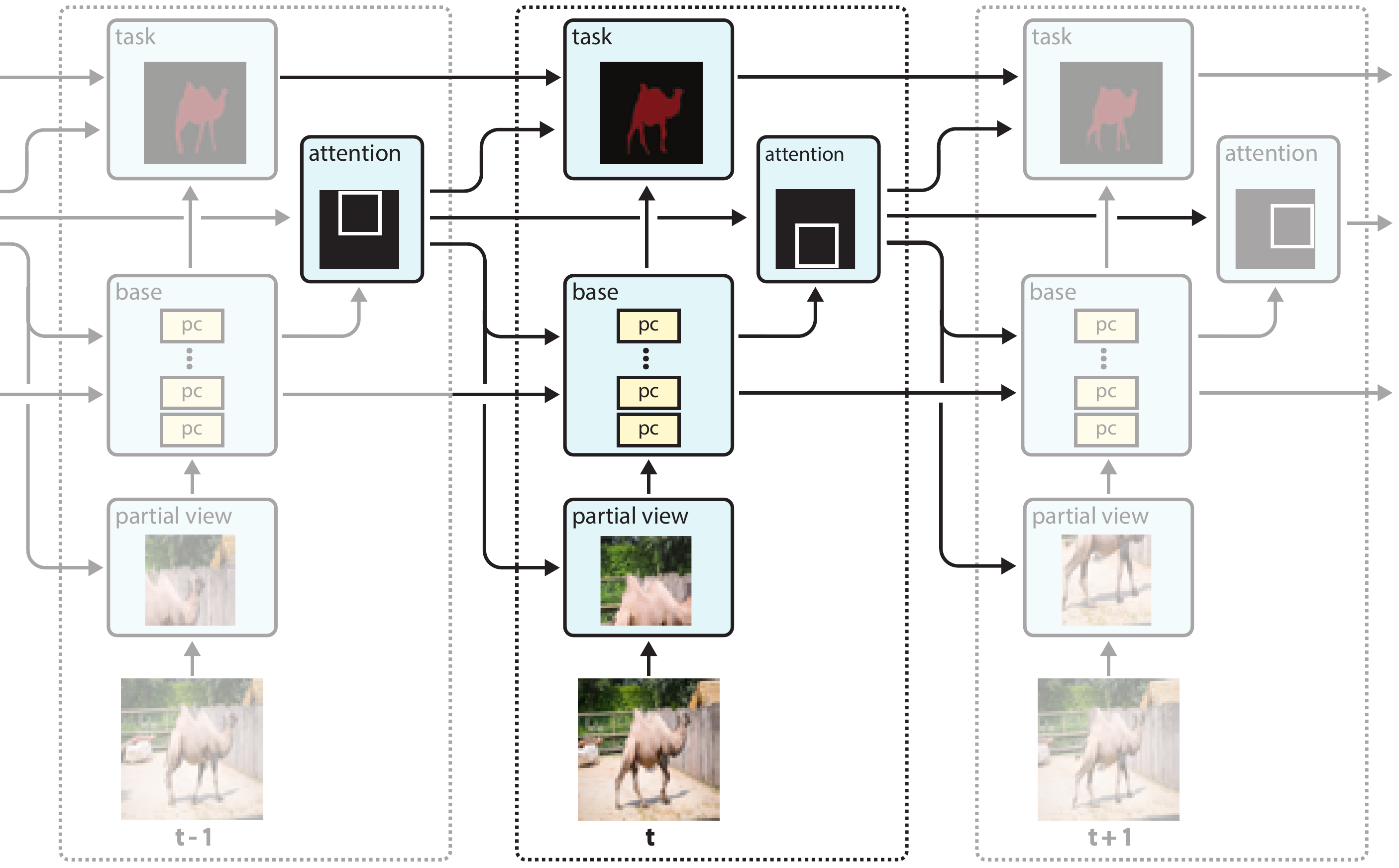}
    \caption{Overview of the Patchwork architecture.\vspace{-0.2cm}}
    \label{fig:teaser}
  \end{subfigure}
  \begin{subfigure}[b]{0.27\textwidth}
    \includegraphics[width=\textwidth]{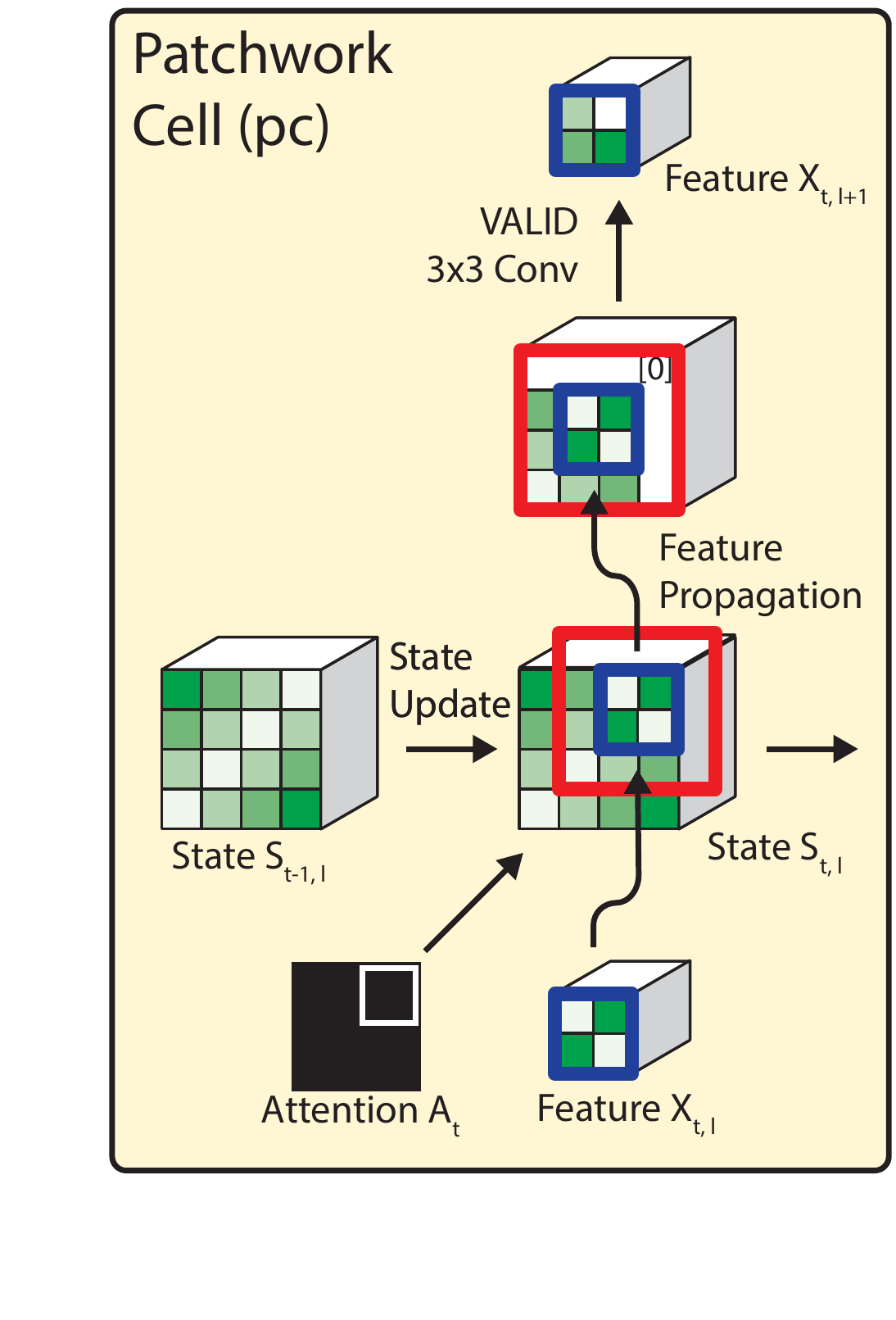}
    \caption{The Patchwork Cell.\vspace{-0.2cm}}
    \label{fig:cell}
  \end{subfigure}
  \caption{
a) The proposed Patchwork architecture. At each time step in the video stream, our method only processes a small sub-window of the frame, but can still reasons about the full frame thanks to a series of stateful Patchwork Cells (pc). b) A zoomed-in view of the stateful Patchwork Cell, which modulates the input feature by encasing it with contextual features from previous states. $t$ denotes the time step while $l$ denotes a particular deep net layer. Please see \secref{sec:patchwork cell} for details. These cells replace all traditional convolutional filters in a network, as shown in \supfigref{fig:mobilenet_full}.
\vspace{-0.5cm}}
\end{figure*}

The human visual system is confronted with an overwhelming amount of information in a constant visual stream. Fortunately, our brain is capable enough to remember our environment from previous experiences and can instantly find the best sequence of visual attention decisions to achieve a seemingly effortless perception of our visual surroundings. As \cite{Carrasco11, Markant14} pointed out, we do not perceive our entire environment at once, and the ``secret'' to our success lies in a clever interplay between our ability to \textit{remember} the bigger picture while \textit{focusing} our attention to the objects of interests.

In recent years, there has been tremendous progress in object detection \cite{Dai16,Girshick15,Liu16,Ren17} and segmentation \cite{Chen18a,Chen18,He17} techniques thanks to the emergence of deep convolutional networks \cite{He16,Krizhevsky12,Simonyan15,Szegedy16}. It is natural to extend these powerful methods to applications that require continuous stream processing. However, our own human visual perception experience suggests that it is not efficient to naively apply single-frame detectors to every frame of a video at a fixed time interval without any temporal context.

Inspired by the human visual attention system \cite{Markant14}, we introduce Patchwork, a model that explores the subtle interplay between memory and attention for efficient video stream processing. \figref{fig:teaser} illustrates an overview of Patchwork. At each time step, Patchwork crops a small window from the input frame and feeds it into the feature extractor network, which has been modified to contain a set of specialized memory cells scattered throughout the body of the network. The network eventually predicts task-specific outputs: a set of boxes for object detection or a mask for segmentation. Besides, the network predicts the attention window for the next frame that is most likely to contain useful information for the task.

The primary motivation for Patchwork is efficient stream processing. In other words, we aim to achieve the highest possible detection or segmentation quality while reducing latency and computational cost. For applications where there is no need to reduce the latency, we can re-invest the saved resources and boost quality. We demonstrate this latency reduction and quality improvement on two benchmark datasets: ImageNet VID \cite{Russakovsky15} for video object detection, and DAVIS \cite{Perazzi16} for video object segmentation. This latency reduction is controlled a priori by a pair of hyperparameters. We explain a few hyperparameter choices in the experimental section. Some choices significantly reduce latency but take a hit in quality, and others save resources but achieve similar quality. There are also configurations where a quality gain can be observed at a comparable amount of computation.

The contributions of this paper are three-fold: 1) We present Patchwork, a recurrent architecture inspired by the human visual perception system to perform efficient video stream processing. Our method makes use of 2) The Patchwork Cell that serves as the memory unit to carry environmental information across time and 3) An attention model that can predict the best location to attend to in the next frame. Our method is trained via Q-learning with novel object detection and segmentation reward functions.

\section{Related work}

\PAR{Efficient stream processing.} Driven by application needs, there is increasingly more interest in running deep learning models fast and in real-time. These advances fall into two buckets. The first bucket contains methods that change the network at its rudimentary level, e.g., quantization \cite{Jacob18, Zhou17}, and layer decomposition \cite{Chollet17,Lebedev14,Sandler18}. The other bucket of methods, where this work also falls into, operates at a higher and algorithmic level. For image object detection, there are SSD \cite{Liu16} and YOLO \cite{Redmon15}, both of which are one-stage detectors aimed to achieve a better speed-accuracy trade-off over the two-stage approaches such as Faster-RCNN \cite{Ren17}. Specifically for video streams, under the assumption that the outputs for neighboring frames are often similar, \cite{Li18, Shelhamer16, Zhu18a} spend heavy computation only on dynamically select key frames, thus trading accuracy for computation. Our work also exploits the temporal consistency, but instead of only selecting key frames, we select spatial locations per frame in the hope of achieving a better latency-accuracy trade-off.

\PAR{Recurrent attention models.} Most attention models so far have been designed to repeatedly look at parts of the same image rather than across different frames in a video. One of the earliest works is the recurrent attention model (RAM) \cite{Mnih14}, which repeatedly places attention windows onto an MNIST  image to classify digits. Several works extended the RAM framework to multi-label classification \cite{Ba15}, image generation \cite{Gregor15} and single-image object detection \cite{Caicedo15}. The object detector in \cite{Caicedo15} gradually moves the attention window around the image until an object is detected, which is vastly different than our approach. More importantly, its complexity grows linearly with the number of objects, where our method has a constant processing time per frame. In the video domain, \cite{Han18, Jayaraman18} applied the attention model to video streams, though they do not address the problem of lost context by the spatial attention, which is a key contribution of this work.

\section{Patchwork}

\begin{figure} \centering
\includegraphics[width=8.2cm]{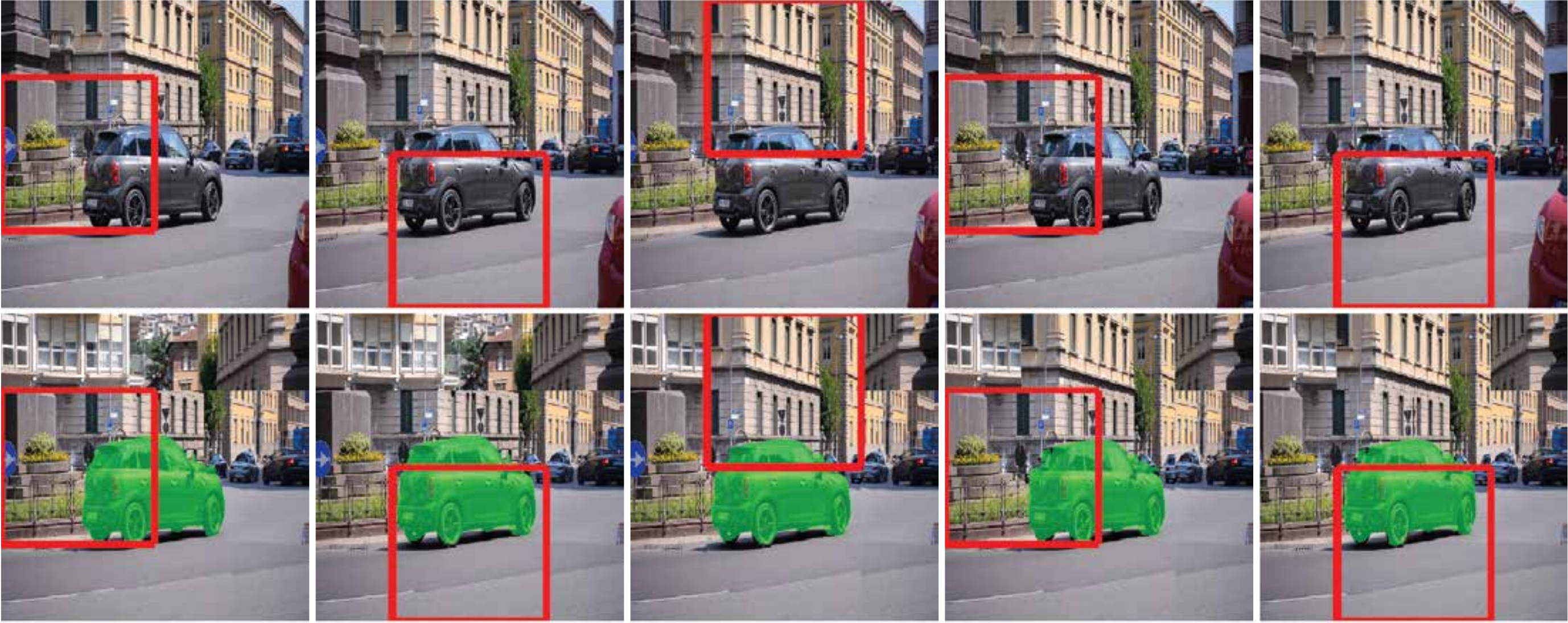}
\caption{
(Best viewed in color) The evolution of a Patchwork Cell memory over five time steps. \textbf{Top}: Input frames with attention windows in red rectangles. \textbf{Bottom}: The Patchwork Cell aggregates raw input patches cropped by the attention windows over time. Note that only features in the red rectangles are updated. Therefore, the feature map looks in disarray, forming a patchwork pattern (\textit{https://en.wikipedia.org/wiki/Patchwork}).
\vspace{-0.5cm}}
\label{fig:samples_2}
\end{figure}

As shown in \figref{fig:teaser}, the Patchwork architecture is a recurrent system where the prediction for the current frame may depend on all previous frames. At each time step, the input frame undergoes four stages: cropping, feature extraction, task-specific prediction (detection or segmentation) and attention prediction. During cropping, a window of fixed size is cropped from the input frame, where the attention predictor from the previous frame dictates the location of the crop. Our choice to limit the window size to be constant is deliberate so that it allows us to control the computational cost ahead of time, as the cost is roughly proportional to the area of the window. For the feature extraction stage, we use a stateful network adapted from a standard backbone network, MobileNetV2 \cite{Sandler18}. We replace \textbf{all} its convolutional layers which have kernel size larger than 1x1 with a custom stateful Patchwork Cell, which appears in details in \secref{sec:patchwork cell}. Finally, the attention and task-specific predictors build on an appropriate layer on top of the feature extraction. Please see \supsecref{sec:Detailed network architectures} for details.

We organize the rest of this section as follows: The two critical pieces of the model, the recurrent attention module and the Patchwork Cell, are described in \secref{sec:recurrent attention} and \secref{sec:patchwork cell}. The last two sub-sections describe the model training: \secref{sec:q-learning reward} describes the reward function for both the object detection and segmentation task, and the rest of the training details are concluded in \secref{sec:training}.

\textit{Patchwork} has two meanings. For one, it is a portmanteau for a PATCH-wise attention netWORK. However, Patchwork is also the English word for a form of needlework where multiple pieces of fabric are sewn together into a larger design---this resembles how the memory within each Patchwork cell appears during inference (see \figref{fig:samples_2}).

\subsection{Recurrent attention}
\label{sec:recurrent attention}

\figref{fig:teaser} shows an overview of the recurrent attention network. In prior literature, the attention window (``attention'' in the figure) is parameterized using its center and size in a continuous space \cite{Ba15, Gregor15, Mnih14}. As for the training, \cite{Gregor15} used a smooth attention window with gradients on the boundary and therefore could train the mechanism end-to-end in a supervised way; \cite{Ba15,Mnih14}, on the other hand, made use of policy gradients, a flavor of reinforcement learning (RL). Note that experiments in these prior works are limited to datasets such as the MNIST and CIFAR, which explains why we had only limited success with either methodology on complex real-world object detection and segmentation tasks. Our best results were obtained from Q-learning on a discrete action space.

This discrete action space consists of all possible attention sub-windows, and is parameterized by two integers $M$ and $N$. $M$ denotes how many times a dimension is sliced, while $N$ denotes how many adjacent slices form an attention window. Our experiments contain three such configurations. For $M=2, N=1$, there are 4 possible windows of relative size $[\frac{1}{2},\frac{1}{2}]$. These windows have the top-left corners at $[\frac{i}{2},\frac{j}{2}], \forall i \in 0,1;j \in 0,1$. $M=4, N=2$ has 9 possible windows of relative size $[\frac{1}{2},\frac{1}{2}]$, with the top-left corners at $[\frac{i}{4},\frac{j}{4}], \forall i \in 0,1,2;j \in 0,1,2$. $M=4, N=1$ has 16 possible windows of relative size $[\frac{1}{4},\frac{1}{4}]$, with the top-left corners at $[\frac{i}{4},\frac{j}{4}], \forall i \in 0,1,2,3;j \in 0,1,2,3$.

Note that $M$ and $N$ control the amount of computation a priori. For example, the $M=2,N=1$ configuration makes each attention window attending to 25\% of the area of a frame, and hence the total computation is reduced to roughly 25\%.

\begin{figure}
\begin{center}
\includegraphics[width=0.4\textwidth]{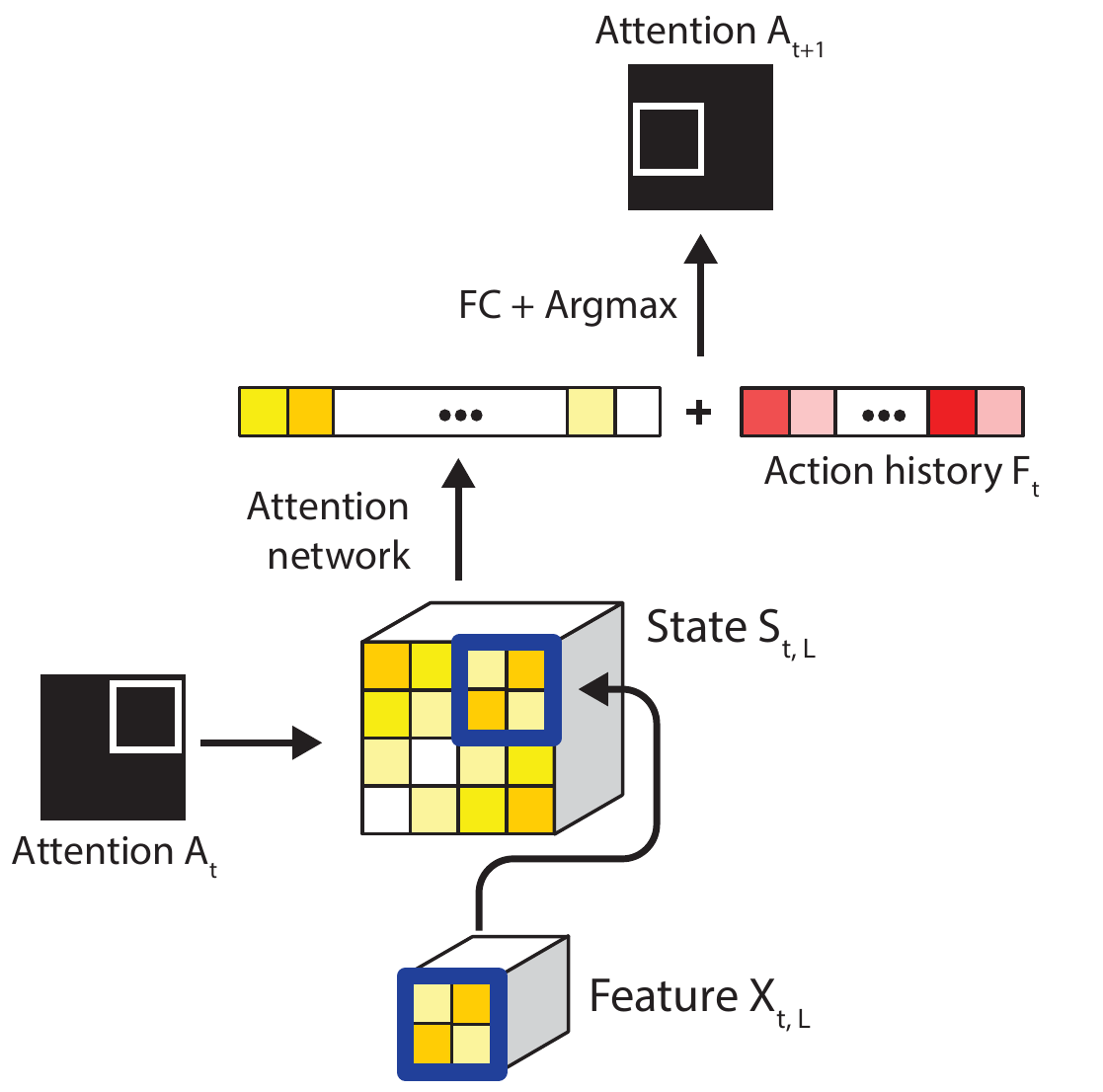}
\caption{
The attention mechanism. The attention network builds on top of the memory unit of a Patchwork Cell and an action history, and predicts the Q-values for the next time step. $t$ denotes the time step and $L$ is a fixed layer in the network. See \secref{sec:recurrent attention} for details.
\vspace{-0.5cm}}
\label{fig:attention_net}
\end{center}
\end{figure}

Next, we construct the attention network that takes a set of features from the network at time $t$ and map them to Q-values $Q(S_t, A_t; \Theta)$ for deep Q-learning (DQN \cite{Mnih15}). The attention network builds on top of the memory unit of a Patchwork Cell, which has a view of the full frame. Additionally, we would also like the attention network to know the action history, to encourage diversity among actions. To this end, we designed a simple exponentially-decaying action history $F_t$: \begin{equation} F_t = \min(\alpha \cdot F_{t-1} + onehot(A_t), 1.0) \end{equation} where $A_t$ is the action (attention window) at the current time step, and $\alpha$ is a discount factor. As shown in \figref{fig:attention_net}, a Patchwork Cell memory is mapped to a one-dimensional feature vector and concatenated with the action history, where a final fully-connected layer is applied to compute the Q-values.

\subsection{Patchwork Cell}
\label{sec:patchwork cell}

Modern deep convolutional networks have receptive fields that are typically larger than the size of the input image. These large receptive fields are crucial in providing later logic (e.g., a detector) with contextual information. However, the hard attention in the recurrent attention module severs these receptive fields, as a cropped input loses any context beyond its own cropped view. Our experiments in \tblref{tbl:abblation_3} show that this loss of context would cause a significant drop in accuracy if not handled properly. To remedy this, we introduce a building block that is the core of this paper, the Patchwork Cell (see \figref{fig:cell}).

The Patchwork Cell is a stateful replacement for the otherwise stateless 2D-convolution layer with the \textit{SAME} padding that appears throughout modern deep neural networks. The 2D-convolution is defined as the mapping $SameConv2d(X, \Theta) \rightarrow Y$, where $X$ is the input feature map, and $\Theta$ is the weight with an odd filter size of $(2k+1)^2$. Without loss of generality, we assume the stride to be 1. Hence, $X_t$ and $Y_t$ have the same spatial resolution [$H$, $W$].

The Patchwork Cell adds two states in the form of a memory cell $S_t$ and an action $A_t$ into the mapping: $Patchwork(X_t, A_t, S_t, \Theta) \rightarrow Y_t$, where the subscript $t$ describes the time step. The cell consists of three components: state update, feature propagation and 2D-convolution with \textit{VALID} padding.

\PAR{State update.} The feature extracted from the sub-window $X_t$ overwrites part of the memory $S_{t-1}$ at locations specified by the initial crop window at the time step $t$, and yields the new state $S_t$:
\begin{equation*}
S^{m,n}_{t} =
\begin{cases}
X^{m - \frac{a_t \cdot H}{h}, n - \frac{b_t \cdot W}{w}}_t, & \text{if $0 \leqslant m - \frac{a_t \cdot H}{h} < H$} \\
&\text{and $0 \leqslant n - \frac{b_t \cdot W}{w} < W$} \\
S^{m, n}_{t-1}, & \text{else}
\end{cases}
\end{equation*}
$\forall m \in 0, 1, ..., \nicefrac{H}{h}, n \in 0, 1, ..., \nicefrac{W}{w}$. The superscripts describe coordinates.  $[a_t, b_t]$ are the relative coordinates of the top-left corner of the attention window. $[h, w]$ are the relative height and width of the window, [0.5, 0.5] and [0.25, 0.25] in our experiments. 

\PAR{Feature propagation.} We modulate the input feature $X_t$ by encasing it with features from the past state $S_{t-1}$. This operation is equivalent to cropping a slightly larger feature $\hat{X}_t$ from the current state $S_t$:

The modulated feature $\hat{X}_t$ 
\begin{equation*}
\hat{X}^{i,j}_t =
\begin{cases}
S^{i-k + \frac{a_t \cdot H}{h}, j-k + \frac{b_t  \cdot W}{w}}_t, & \text{else if $0 \leqslant i-k + \frac{a_t \cdot H}{h} < \frac{H}{h}$} \\
&\text{and $0 \leqslant j-k + \frac{b_t  \cdot W}{w} < \frac{W}{w}$} \\
0, & \text{else}
\end{cases}
\end{equation*}
$\forall i \in 0, 1, ..., H + 2k, j \in 0, 1, ..., W + 2k$.

Finally, $Y_t$ is obtained by performing a 2D-convolution with \textit{VALID} padding on the modulated feature $\hat{X}_t$: $Y_t = ValidConv2d(\hat{X}_t, \Theta)$.

The Patchwork Cell is placed numerous times throughout the network, namely, in front of every \textit{SAME} padding convolution with a kernel size larger than 1, and before global operations. An exact comparison between the off-the-shelf detector and segmenter and their respected Patchwork-modified versions appears in \supsecref{sec:Detailed network architectures}.

Note that the context provided by the Patchwork Cells is not exact and is an approximation in two ways: 1) One type of approximation is the temporal staleness in the cell memory, as these stale features are extracted from past frames and used to augment features of the current frame. This approximation error becomes more severe the more drastically a scene changes. 2) The context gradually provided by the series of Patchwork Cells is not numerically identical to a natural receptive field, i.e., that of the original pre-cropped input frame. However, in a toy experiment, we found that this type of error is almost negligible, as shown in \supsecref{sec:Patchwork Cell approximation}.

\begin{figure*}[!htbp]
  \centering
\includegraphics[width=17.5cm]{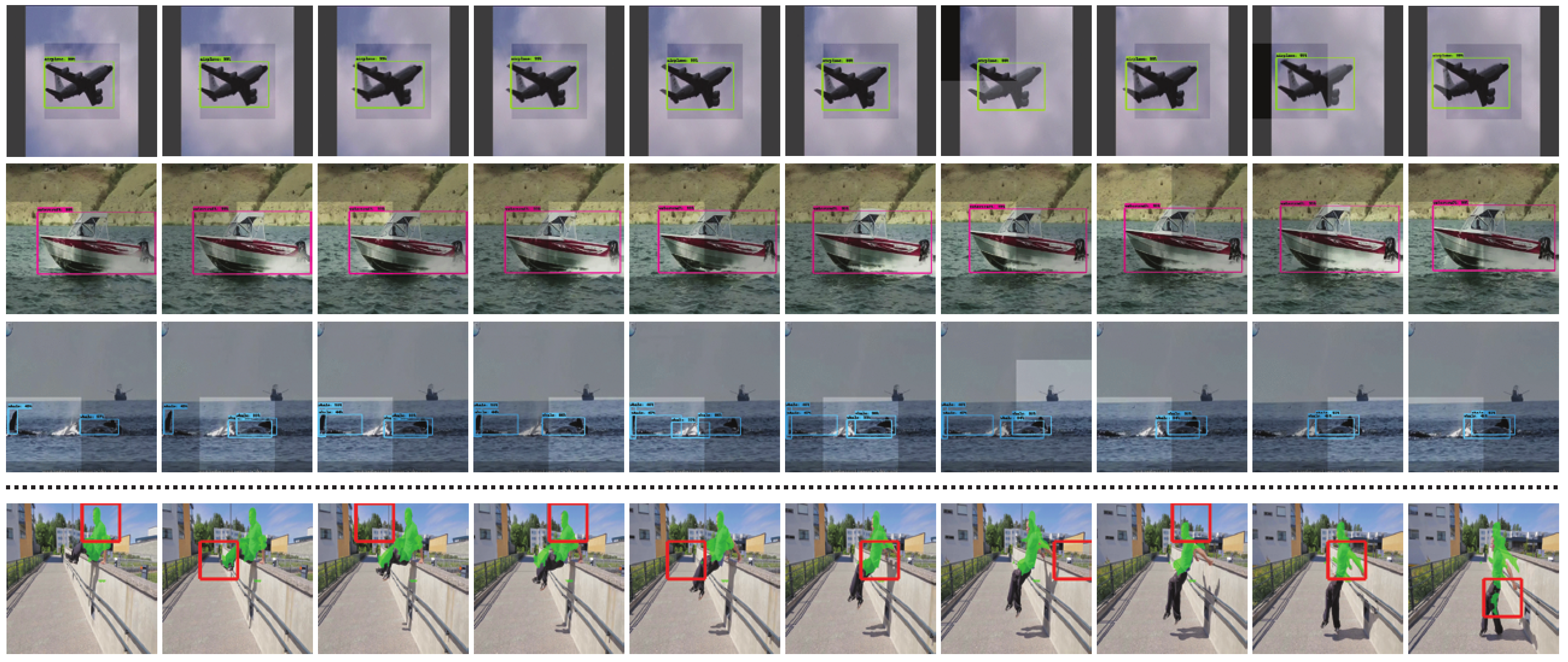}
\caption{
(Best viewed in color) Sample results for the Patchwork network over ten consecutive frames. \textbf{First three rows}: object detection results on ImageNet VID. Attention window appears in darker or lighter shading. \textbf{Top row}: When the object does not move and can be enclosed within one possible attention window, the window stays until the object away. \textbf{2nd row}: When the object is too large for one attention window, the attention window moves around to cover the whole object. \textbf{3rd row}: When there are multiple objects of interest spread far apart, the attention window jumps between these objects. \textbf{Bottom row}: object segmentation results on DAVIS 2016. Green denotes the segmentation mask, and the red rectangle shows the attention window. Note that the attention window does not simply follow the object of interest. The Q-learning enables a non-greedy attention mechanism that focuses on smaller and potentially new parts while memorizing parts that have been attended before.
\vspace{-0.5cm}}
\label{fig:samples}
\end{figure*}

\subsection{Q-Learning reward}
\label{sec:q-learning reward}

Now that we have defined the state, the action, and the Q-function, we need to choose an appropriate reward to train the attention network. A good time-difference ($TD(0)$) reward \cite{Sutton98} should be continuous, as a moderately good action shall receive a smaller reward than a vastly superior action. It should also have low variance, and the reward should associate more with the action than the overall difficulty of the scene. Hence, it is inappropriate to naively use the mean-average-precision (mAP) for object detection or the mean intersect-over-union (mIoU) for segmentation as the reward function, as they largely depend on the difficulty of the scene.

In the literature, subtracting a fixed baseline from the raw reward reduces the variance \cite{Mnih14}. For our video case, we can do slightly better by using the prediction of the previous frame to form the baseline for the current action. Namely, we define the $TD(0)$ reward $R_t$ as:
\begin{equation}
R_t = \max(0, f(gt_t, p_t) - f(gt_t, p_{t-1}))
\end{equation}
where $gt$ and $p$ are groundtruth and predictions, $f(gt, p)$ is the metric, and $t$ is the time step. The intuition here is that a reward should only be given if the new attention window has contributed to an increase of a metric $f$ over doing nothing. We force the reward to be non-negative since any negative reward would indicate that having a fresh view of the scene is detrimental in explaining a scene, which should be strictly due to noise.

As for the task-specific metric $f$, for segmentation, we use the frame-wide mIoU metric (also known as the J-measure in DAVIS \cite{Perazzi16}). However, for object detection, the mAP measure for object detection is inappropriate as it is quantized and does not reflect incremental improvements. Instead, we draw inspiration from the segmentation task, and define an average box overlap measure as the reward function for object detection: \begin{equation} f(gt, p) = \frac{1}{K} \sum_{1}^{K} IoU(gt_k, \tilde{p_k}) \end{equation} where each prediction with a score larger than 0.5 is greedily matched to one of the groundtruth boxes $gt_k$ to yield $\tilde{p_k}$. We then average the IoU scores over all $K$ groundtruth boxes. Note that we omit the class label altogether for simplicity.

We use the DDQN \cite{vanHasselt16} method to train the weights $\Theta$ in the Q-function approximation $Q(S_t, A_t; \Theta)$:
\begin{equation}
  \Theta \leftarrow \Theta + \alpha (Y^Q_t - Q(S_t, A_t; \Theta)) \nabla_{\Theta} Q(S_t, A_t; \Theta)
\end{equation}
where
\begin{equation}
  Y^Q_t \equiv R_{t+1} + \gamma Q(S_{t+1}, \operatorname*{arg\,max}_a Q(S_{t+1},
  a; \Theta); \Theta')
\end{equation}
$\Theta'$ is a delayed copy of the weight to make the Q-learning procedure less overly confident. $\gamma$ is the discount factor and $\alpha$ is the learning rate.

\subsection{Training details}
\label{sec:training}

There are two significant challenges associated with the training of Patchwork. First, the amount of video training data in the benchmark datasets in the experiments is tiny. ImageNet VID has roughly a thousand unique training videos (technically 3862 but not all unique) whereas DAVIS 2016 only has 30. We decided to make use of the vast amount of single-frame data sources by generating fake videos from them. For a single static image, we randomly sample 2 boxes and build a video sequence by moving the view from one box to the other. We make sure that at least one of the two boxes overlaps with a certain amount of groundtruth detection boxes or segmentation foreground pixels, so that the faked video is not all background. The speed with which the view window moves adheres to a Gaussian distribution. For both the detection and segmentation tasks, we use the COCO and Pascal VOC datasets to augment the original training sets. The usage of these datasets to aid ImageNet VID is not new, as \cite{Liu18} also used them to pre-train their networks.

A second challenge is training with highly correlated samples in mini-batches. This correlation makes the optimization less efficient since multiple samples push gradients roughly in the same direction. More importantly, this makes batch normalization (BN) \cite{Ioffe15}, a critical module to train any deep image model, no longer applicable. To remedy this, we adopt a three-stage training procedure: 1) The video data is split into single frames and mixed with augmentation data to train a single-image model with BN; 2) We place the trained single-image model in a recurrent Patchwork setting and only train the attention network part with Q-learning; 3) With BN frozen, we fine-tune  the main network jointly using the task-specific loss and the attention network with Q-learning at a lower learning rate.

\section{Experiments}
\label{sec:experiments}

Before diving into the experimental results, we would like to discuss one of the core metrics that is the per-frame latency of the system, which has had different definitions in the past \cite{Liu18, Sandler18}.

\PAR{Average vs. maximum latency.} Some algorithms require a variable amount of computation per frame \cite{Feichtenhofer17, Zhu18}. One notable case is the naive application of a slow model every $K$ frames (keyframes) in a sequence to achieve real-time. Predictions for non-key frames copy from the closest past keyframe. Since copying does not incur any real latency, $K - 1$ frames have no latency at all, while one single keyframe takes $T$ amount of time. In this case, the average latency is $\nicefrac{T}{K}$, which is very different than the maximum latency at $T$. Both the mean and maximum metrics have their own merits. However, we argue that for most latency-sensitive application, the maximum latency is more critical, as the peak device usage is the bottleneck of the system. Therefore, we report the maximum latency in the experiments section. The average latency appears in \supsecref{sec:Detailed results}.

\PAR{Theoretical (FLOPs) vs. empirical latency metric (seconds).} The theoretical latency metric measures the latency in the number of floating point operations, which represents the intrinsic complexity of an algorithm. It is independent of the particular implementation and hardware, and therefore, conclusions drawn based on the theoretical measure may remain significant for a long time as implementations are optimized and hardware gets deprecated. The empirical latency metric, on the other hand, measures the latency in absolute time units (e.g., milliseconds). It depends on a particular implementation, so a conclusion can be valid for one hardware, but does not hold for another. We report both metrics in the experiments section as complete as possible. The empirical latency in our experiments is measured using Tensorflow and on a single core Intel Xeon at 3.7GHz. 

\PAR{Model variants.} We conducted experiments on multiple variants of the baseline single-frame model and Patchwork for both the object detection and segmentation tasks. The variations include: \textbf{depth}: Depth multiplier, with which the number of channels in all layers is multiplied. The latency is roughly proportional to the square of this multiplier. \textbf{flip}: Also process the left-right flipped frame and average predictions, and doubles the latency. \textbf{resolution}: The default resolution is 384x384. 0.25 resolution has roughly 25\% of the original latency. \textbf{interval}: The model runs on keyframes that are this many time steps apart, and predictions from keyframes propagate to non-keyframes. While the average latency is proportional to the reciprocal interval, the interval alone has no impact on the maximum latency. \textbf{delay}: Predictions are delayed by this many time steps. The delay has no impact on the average latency but can reduce the maximum latency at when combined with a well-chosen interval.

\subsection{Object detection}
\label{sec:Object detection}

\begin{figure*}[!htbp]
  \centering
\includegraphics[width=17.5cm]{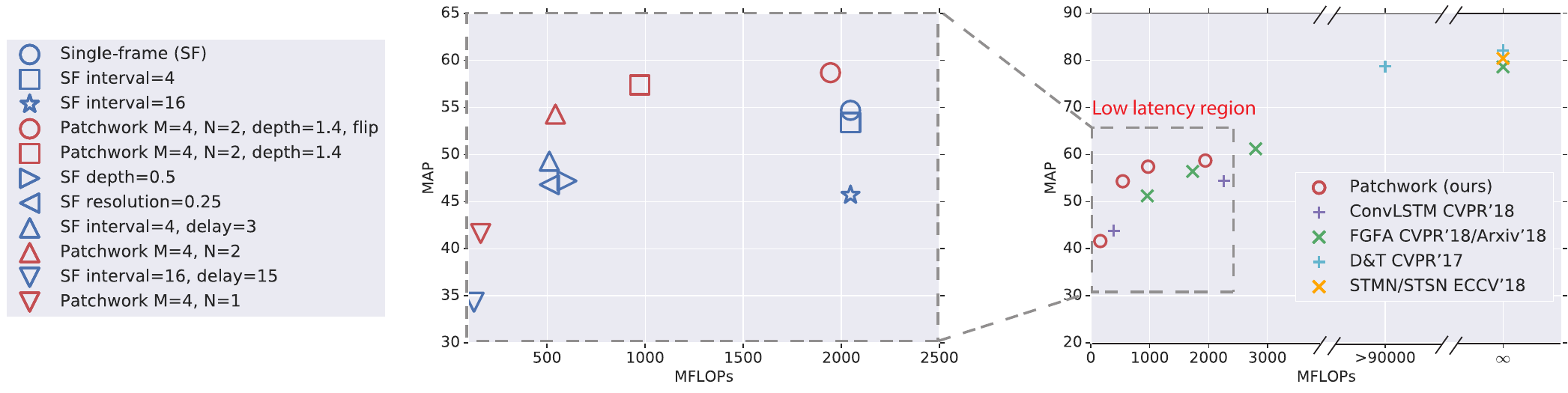}
\caption{
(Best viewed in color) A theoretical maximum latency vs. accuracy comparison on ImageNet VID. \textbf{Left figure}: Comparison between single-frame and Patchwork variants (exact values given in \suptblref{tbl:det_results}). \textbf{Right figure}: We put Patchwork in perspective with other published works on ImageNet VID: ConvLSTM \cite{Liu18}, FGFA \cite{Zhu18, Zhu18a}, D\&T \cite{Feichtenhofer17}, STSN \cite{Bertasius18} and STMN \cite{Xiao18}. Most of these use the ResNet-101 as the feature extractor, which by itself has 90 billion FLOPs when applied on a 600x600 image as described in the papers. Methods that require to look ahead for more than ten frames or optimize for the whole video at once are considered to have an infinite latency. 
\vspace{-0.5cm}}
\label{fig:det_results}
\end{figure*}

\begin{figure}
  \centering
\includegraphics[width=0.35\textwidth]{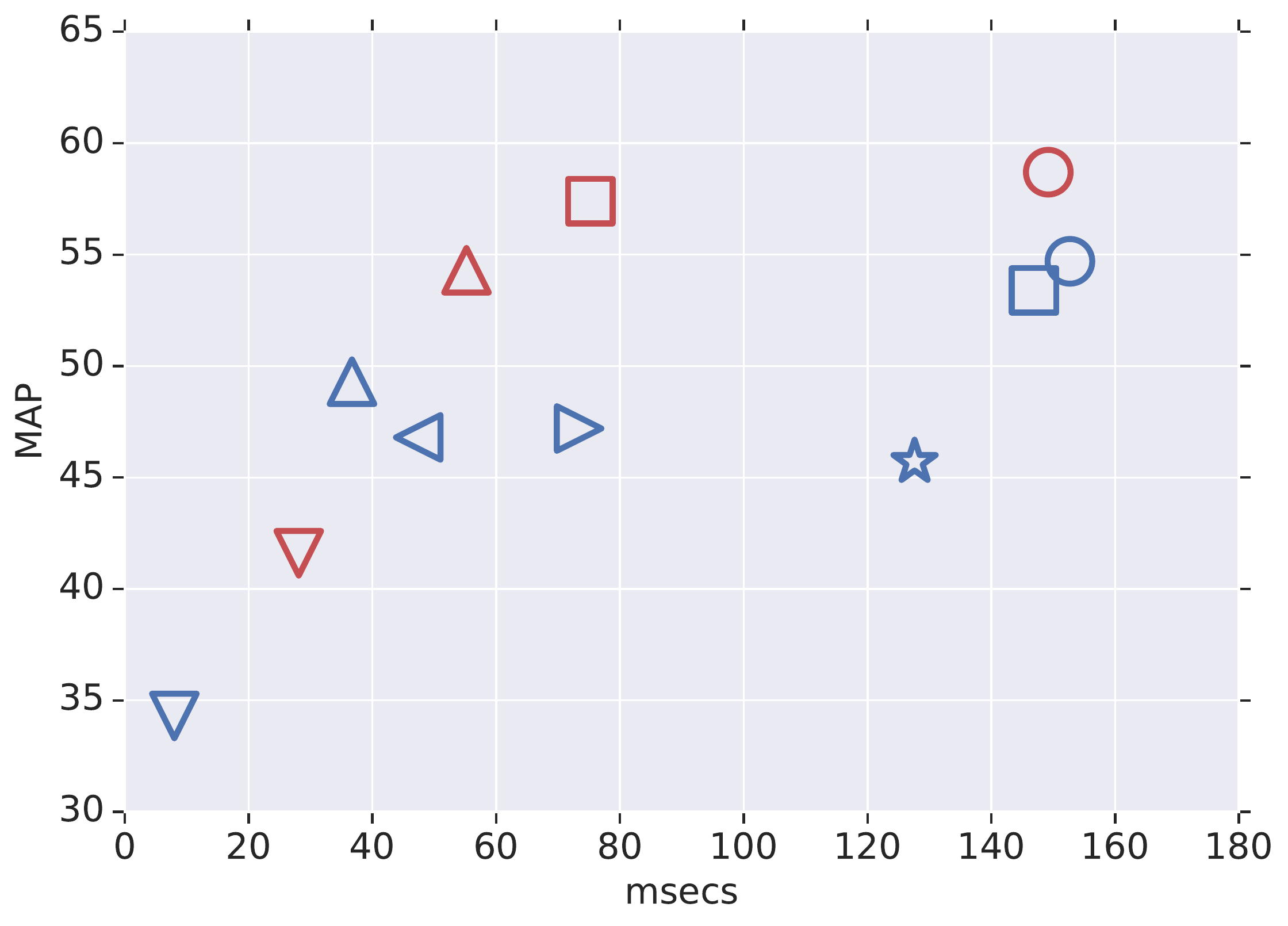}
\caption{
An empirical maximum latency vs. accuracy comparison between single-frame (blue) and Patchwork (red) variants. Same legends as in \figref{fig:det_results}.
\vspace{-0.5cm}}
\label{fig:det_results_msecs}
\end{figure}

For the object detection task, we conducted our experiments on ImageNet VID 2015 \cite{Russakovsky15}, which contains 30 categories with 3862 training and 555 validation video clips. We report our results on the validation set and the mean-average-precision (mAP) at 0.5 IoU, as it is the norm for this dataset.

The system inherits from an off-the-shelf MobileNetV2 + SSD detection
architecture \cite{Sandler18}. In this work, we do not yet consider more expensive models, e.g., ResNet + Faster RCNN, since these models are too slow to run in real-time. For the baseline, we trained a single-frame detection model by fine-tuning from an ImageNet checkpoint using data sampled from ImageNet VID, ImageNet DET and COCO. This baseline detector achieves 54.7\%, which is competitive with other recent works that focus on the low-latency case \cite{Liu18, Zhu18a}.

The left figure in \figref{fig:det_results} shows a theoretical maximum latency vs. accuracy
comparison between Patchwork variants and their equivalent single-frame setups.
Patchwork is consistently superior than the baseline. Most notably, the M=4,N=2
configuration at roughly 25\% of its original computation, only loses 0.4\% in
mAP (54.7\% vs. 54.3\%) compared to the baseline single-frame MobileNetV2 model.
Moreover, when we reinvest the saved resource in additional channels in the
layers and model ensembling, we show that we can boost the results to 58.7\% at
the original computational cost. The empirical maximum latency vs. accuracy figure tells a
similar story (\figref{fig:det_results_msecs}) in the high and mid-accuracy
range. For the low latency range, the overhead caused by the Patchwork ops
starts to dominate. This overhead is mostly because the Patchwork ops are not as
optimized as the rest of the network, which mostly consists of highly optimized
convolutions. Sampled qualitative results appear in \figref{fig:samples}.

Two ablations studies were conducted to highlight the two main contributions of this work: the Q-learned attention mechanism and the Patchwork cell.

\begin{table}
\centering
\begin{tabular}{cclcc}
\hline
  M & N & Policy & MFLOPs & mAP $\uparrow$ \\
\hline
- & - & Single-frame & 512 & 49.3 \\ 
4 & 2 & Random & 541 & 52.5 \\ 
2 & 1 & Scanning & 541 & 51.7 \\ 
  4 & 2 & DQN & 543 & \textbf{54.3} \\ 
\hline
- & - & Single-frame & 128 & 34.3 \\ 
4 & 1 & Random & 161 & 37.0 \\
4 & 1 & Scanning & 161 & 37.9 \\ 
  4 & 1 & DQN & 162 & \textbf{41.6} \\ 
\hline
\end{tabular}
\caption{
A comparison between different attention policies. Single-frame and DQN results match those in \figref{fig:det_results}. \textbf{Random} picks an action at random. \textbf{Scanning} iterates over the action space following a fixed pattern.
}
\label{tbl:abblation_2}
\end{table}

\begin{table}
\centering
\begin{tabular}{lcc}
\hline
  Policy & Patchwork Cell? & mAP $\uparrow$ \\
\hline
  \multirow{2}{*}{Scanning} & \ding{55} & 39.5 \\ 
  & \ding{51} & 51.7 \\ 
\hline
  \multirow{2}{*}{DQN} & \ding{55}  & 43.2 \\ 
  & \ding{51} & 54.3 \\ 
\hline
\end{tabular}
\caption{An ablation study demonstrating the necessity for the proposed Patchwork Cell in recurrent attention models \cite{Mnih14}. Experiments are performed using both the reinforced and the scanning policy.
\vspace{-0.5cm}}
\label{tbl:abblation_3}
\end{table}

\begin{figure*}[!htbp]
\centering
\includegraphics[width=17.5cm]{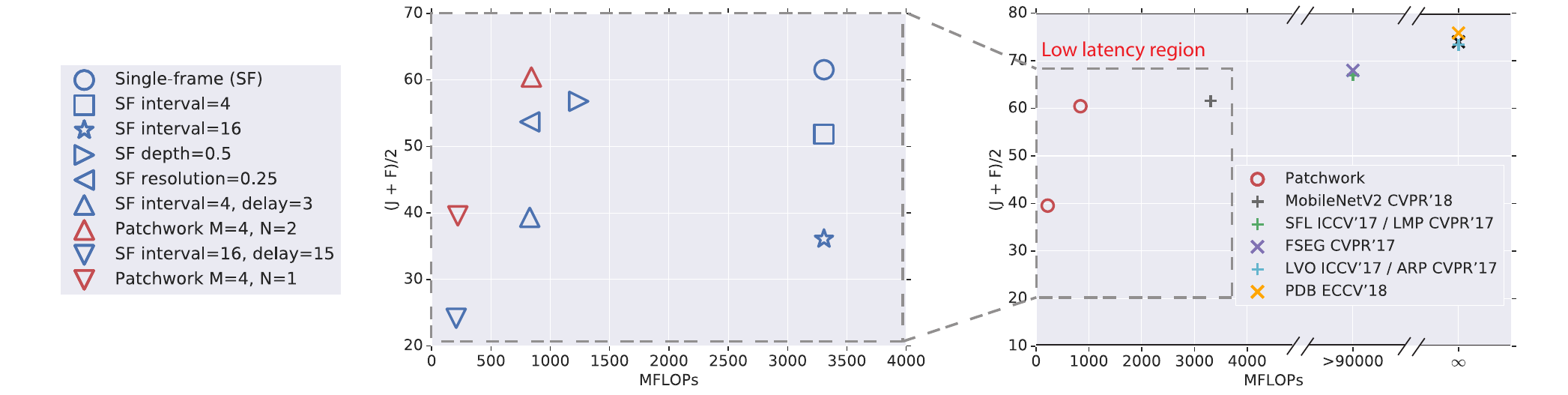}
\caption{
(Best viewed in color) A theoretical maximum latency vs. accuracy comparison on the DAVIS 2016 video object segmentation dataset. $\mathcal{J}$ and $\mathcal{F}$ are the metrics commonly used in DAVIS. \textbf{Left figure}: Comparison between single-frame and Patchwork variants (exact values given in \suptblref{tbl:seg_results}). \textbf{Right figure}: We put Patchwork in perspective with other published works on DAVIS 2016: MobileNetV2 \cite{Sandler18}, SFL \cite{Cheng17}, LMP \cite{Tokmakov17}, FSEG \cite{Jain17}, LVO \cite{Tokmakov17a}, ARP \cite{Koh17} and PDB \cite{Song18}. Most works are based on ResNet-101 and has a substantially higher latency than Patchwork.
\vspace{-0.2cm}}
\label{fig:seg_results}
\end{figure*}

\begin{figure}
  \centering
\includegraphics[width=0.35\textwidth]{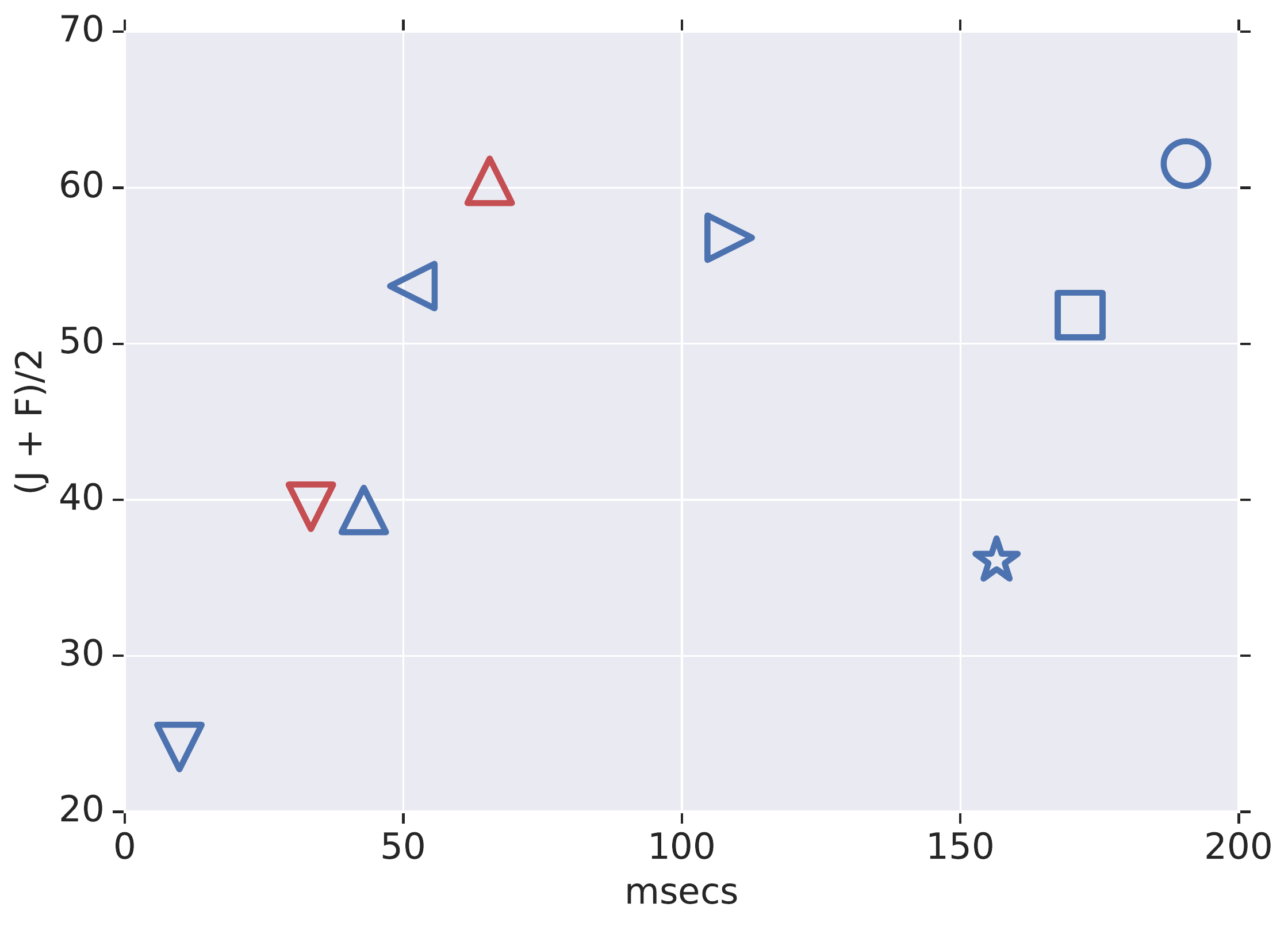}
\caption{
An empirical maximum latency vs. accuracy comparison between single-frame (blue) and Patchwork (red) variants. Same legends as in \figref{fig:seg_results}.
\vspace{-0.2cm}}
\label{fig:seg_results_msecs}
\end{figure}

\PAR{Q-learned vs. hand-designed policies.} Instead of training the attention mechanism via Q-learning, one could use a manually designed policy. One such policy is the random policy, where we select one of the possible actions at random. Another baseline policy is the scanning policy, where the policy deterministically scans through all possible spatial locations. \tblref{tbl:abblation_2} shows a comparison on the \nicefrac{1}{4} and the \nicefrac{1}{16} cost setups. In both cases, the results from the random and scanning policies are fairly comparable. The Q-learned policy is roughly 2-4\% better than the manually designed policies, which are in turn 2-3\% better than the best baseline single-frame method with a comparable maximum latency. In other words, even without a smart attention mechanism, one can still achieve a reasonable improvement by using Patchwork Cells alone.

\PAR{Patchwork Cell vs. vanilla recurrent attention network.} The Patchwork Cell is a core contribution to previous recurrent attention models \cite{Mnih14} and is the key to make them successful on complex object detection and segmentation tasks. \tblref{tbl:abblation_3} shows the results of the Patchwork network with or without the Patchwork Cells, by using either the scanning policy or the DQN-learned policy. In both cases, we observe a significant drop in mAP if the Patchwork Cells are not present.

Finally, the right figure in \figref{fig:det_results} shows our results with other recent publications on ImageNet Video. Note that most of the other works are based on ResNet-101 and has a significantly higher latency than Patchwork and are not suited for latency-sensitive applications.

\subsection{Object segmentation}
\label{sec:Object segmentation}

DAVIS \cite{Perazzi16} has been instrumental in driving the research in video object segmentation. It has been used in two setups: in the semi-supervised setup where the groundtruth segmentation mask for the first frame is present, or the unsupervised configuration where there is no prior information at all. For the 2016 version of DAVIS, where there is at most one object per video, the unsupervised setting becomes a binary foreground-background segmentation task. For simplicity, we choose to test Patchwork in the unsupervised environment. Note that the task of segmenting the foreground per frame is relatively ambiguous and DAVIS is not exhaustively labeled. However, we hope that an algorithmic improvement is still reflected in an increase in our quantitative measure. DAVIS 2016 contains 30 training and 20 validation sequences, each of which includes a single object, be it a person on a bike, a car in a busy street.

Our baseline single-frame architecture is again directly taken from the MobileNetV2 paper \cite{Sandler18}. A maximum latency vs. accuracy
comparison is shown in the left figure in \figref{fig:seg_results} and in
\figref{fig:seg_results_msecs} (see \supsecref{sec:Detailed results} for more details). Patchwork occupies the Pareto points in the mid- and low latency range. Notably, at roughly \nicefrac{1}{4} of computation, we only observe a minimal loss in the $\mathcal{J}$ (63.6\% vs. 62.1\%) and the $\mathcal{F}$ (59.5\% vs. 58.8\%) metrics. Results from other works on the unsupervised DAVIS 2016 appears in the right figure in \figref{fig:seg_results}. Note that most referenced works are based on ResNet-101 and has a substantially higher latency than Patchwork.

\section{Conclusion}

The lost spatial context presents a fundamental limitation for the hard attention mechanism in deep networks, and we believe it is one of the main reasons why the hard attention idea has not been more popular in real-world applications. As demonstrated in the experiments, Patchwork successfully mitigates this issue. Together with a Q-learned attention mechanism, in the Patchwork architecture, we present a new paradigm to efficient stream processing.

%% file: supp_content.tex
\appendix

\section{Detailed network architectures}
\label{sec:Detailed network architectures}


\begin{figure}[b]
  \begin{subfigure}[b]{0.21\textwidth}
    \includegraphics[width=\textwidth]{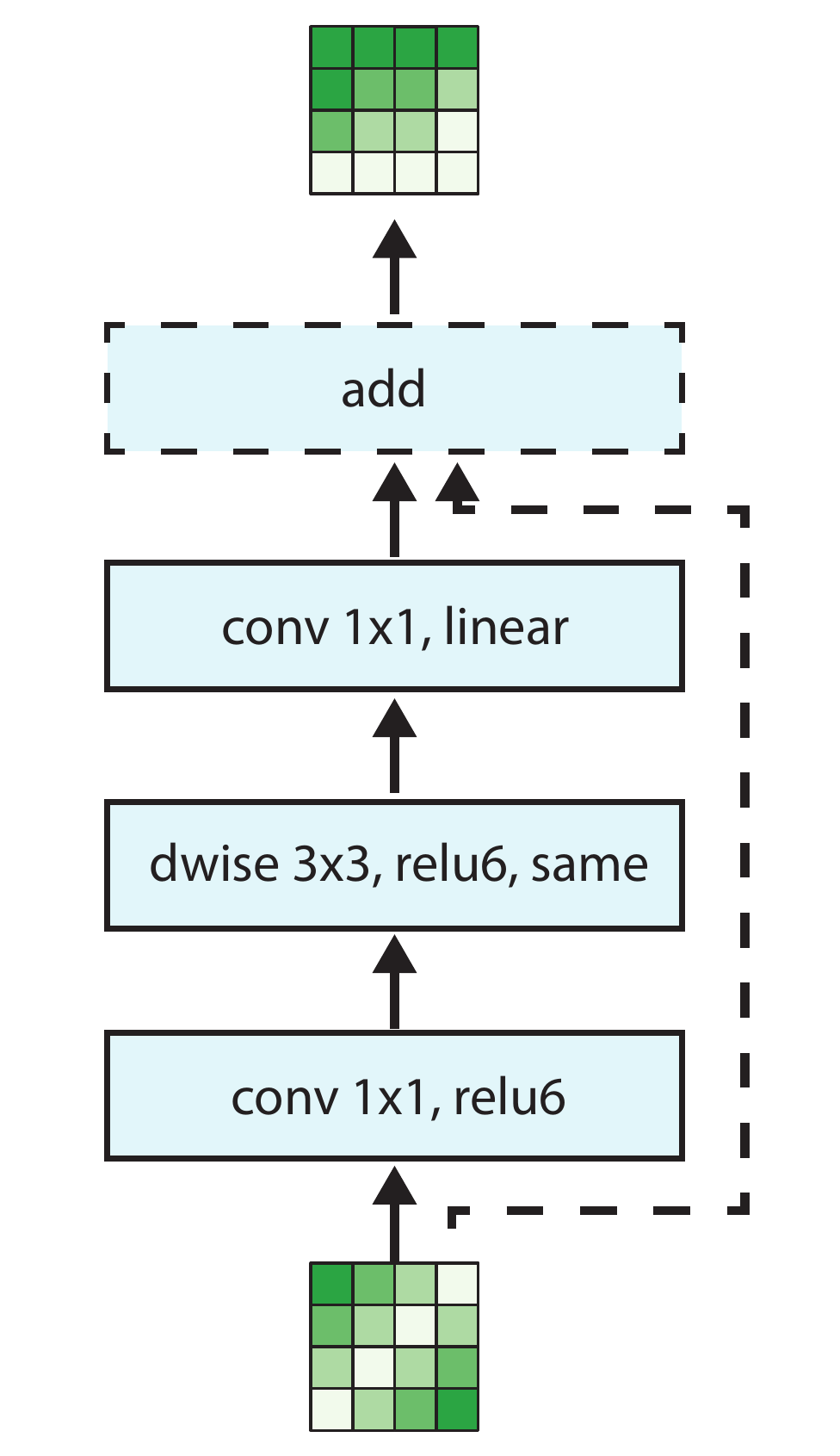}
    \caption{}
    \label{fig:mobilenet}
  \end{subfigure}
  \begin{subfigure}[b]{0.27\textwidth}
    \includegraphics[width=\textwidth]{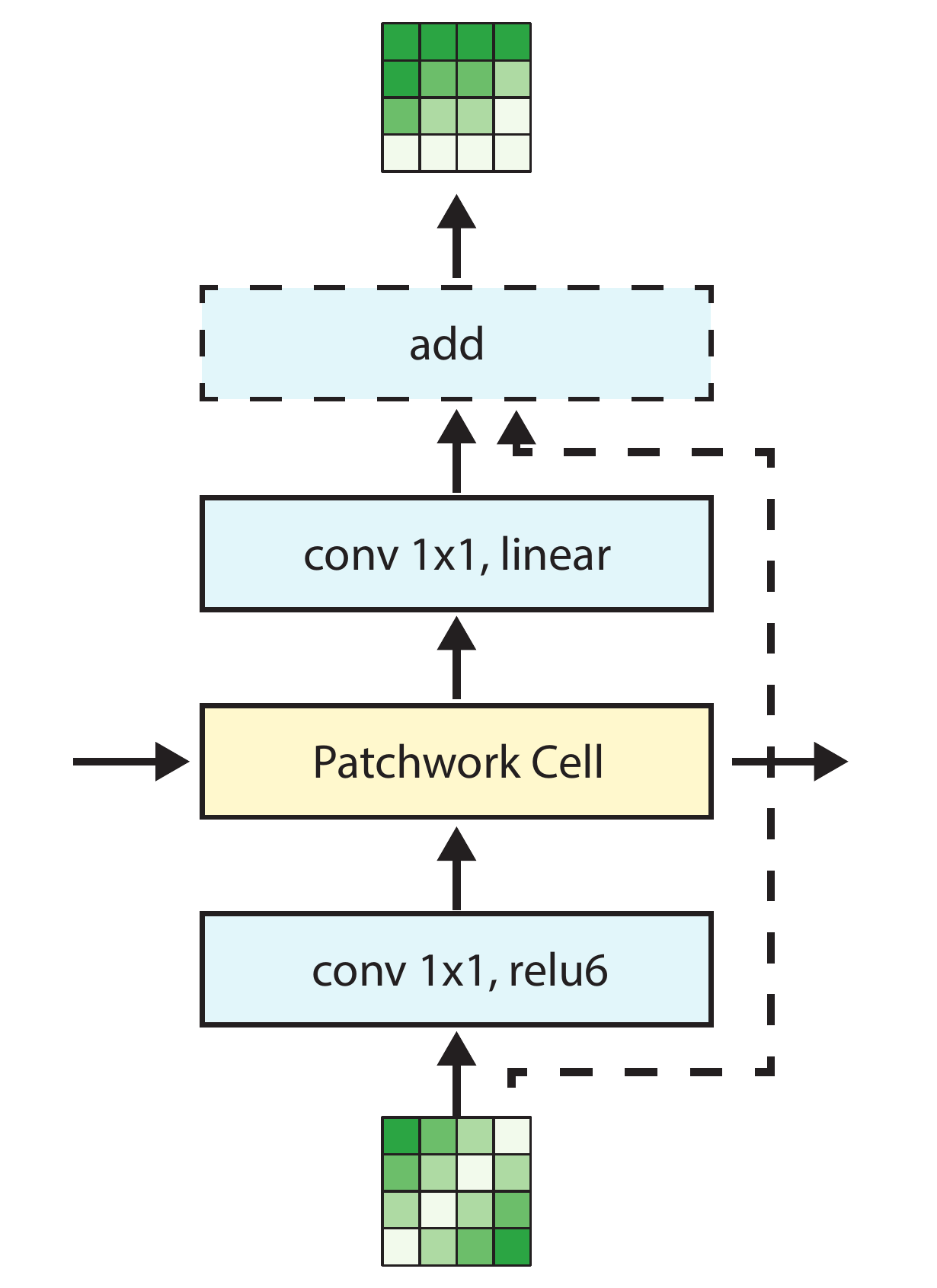}
    \caption{}
    \label{fig:mobilenet_2}
  \end{subfigure}
  \caption{
a): A regular MobileNetV2 block \cite{Sandler18}. The skip connection and the ``add'' op (dotted) are present only when the stride is 1. b): A modified MobileNetV2 block that includes a stateful Patchwork Cell as described in \secref{sec:patchwork cell}.
}
\label{fig:mobilenet_full}
\end{figure}


Here, we discuss in details the network architecture. All Patchwork experiments use the MobileNetV2 \cite{Sandler18} backbone, which consists of bottleneck blocks as shown in \figref{fig:mobilenet}. One bottleneck block consists of a 1x1 convolution to expand the number of channels, a 3x3 separable convolution layer with an optional stride, and a 1x1 convolution to project the number of channels back. \figref{fig:mobilenet_2} shows the modified MobileNetV2 block, where we replace the separable convolution with the \textit{SAME} padding with a stateful Patchwork Cell that contains a separable convolution with the \textit{VALID} padding.

We concatenate such stateful blocks together to form our stateful MobileNetV2 backbone network. An exact description appears in in \tblref{tbl:mobilenet_architecture}.

\begin{figure}
  \centering
  \includegraphics[width=7.2cm]{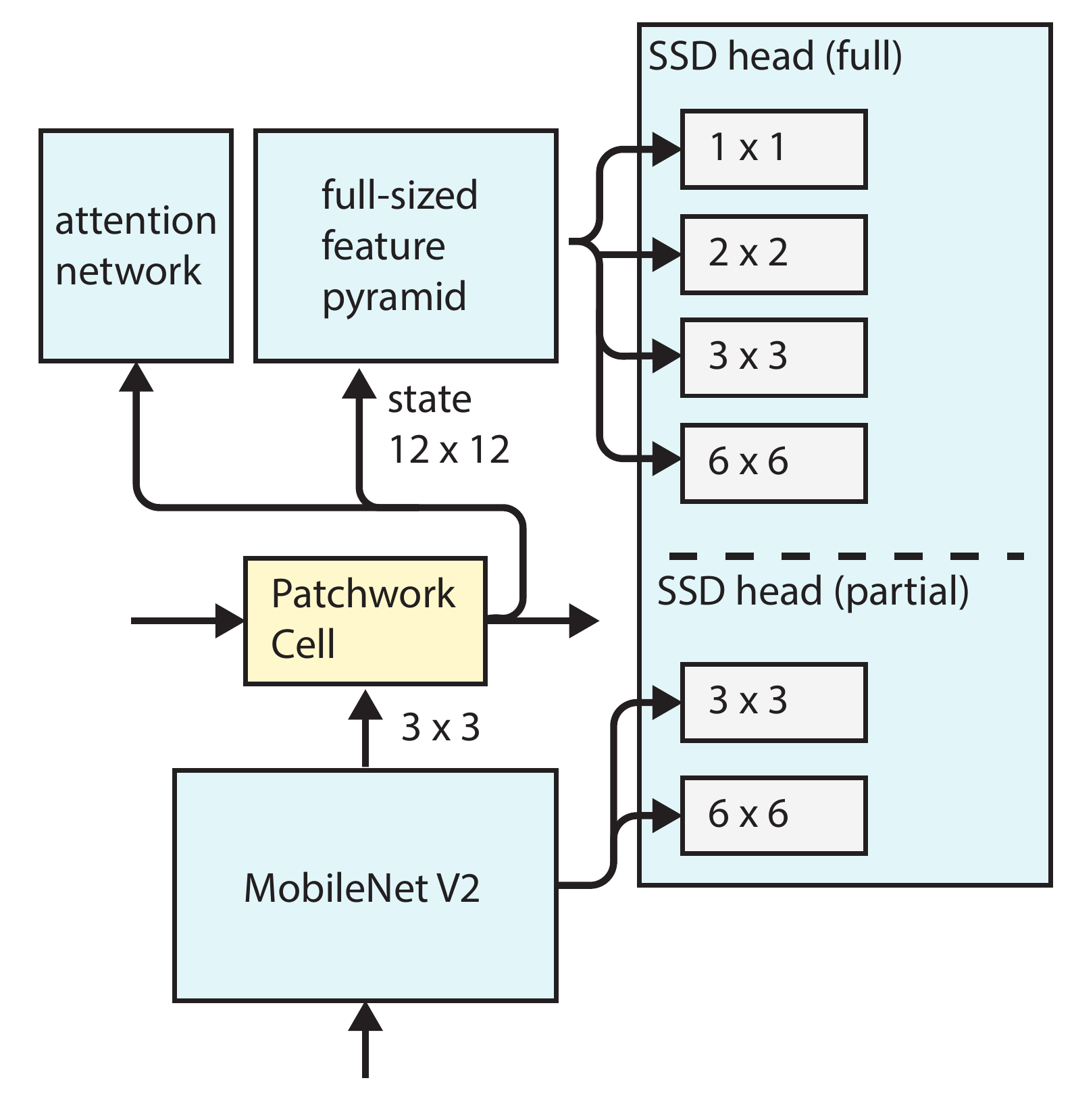}
\caption{
A detailed look at the Patchwork object detector head. High-resolution SSD \cite{Liu16} heads operate on the cropped feature maps, and only produces predictions for the partial window. A Patchwork cell restores a full feature map from its memory, on top of which we build a feature pyramid. Low-resolution SSD heads then operate on this feature pyramid.
\vspace{-0.5cm}}
\label{fig:detector}
\end{figure}

For the object detection task in \secref{sec:Object detection}, the detector head builds on top of the modified stateful MobileNetV2, as shown in \figref{fig:detector}. Smaller objects are predicted by the high-resolution SSD heads that operate directly on the partial feature map. The memory of a Patchwork Cell restores a feature map for the full-sized frame, which allows the prediction of large objects within the SSD \cite{Liu16} framework. The feature pyramid consists of 4 layers of separable convolutions, which gradually reduce the spatial resolution.  The attention network, which consists of two convolution layers and one fully-connected layer also builds on top of the restored full-sized feature map.

The segmenter head in \secref{sec:Object segmentation} builds on top of the modified stateful MobileNetV2 backbone. Since the pixel-wise segmentation operates locally, it does not require a restored full-frame-sized feature map.

\begin{table}
\centering
\begin{tabular}{llcccc}
\hline
  Input & Operator & t & c & n & s \\
\hline
  $96^2 \times 3$ & s-conv2d & - & 32 & 1 & 2 \\
  $48^2 \times 32$ & s-bottleneck & 1 & 16 & 1 & 1 \\
  $48^2 \times 16$ & s-bottleneck & 6 & 24 & 2 & 2 \\
  $24^2 \times 16$ & s-bottleneck & 6 & 32 & 3 & 2 \\
  $12^2 \times 32$ & s-bottleneck & 6 & 64 & 4 & 2 \\
  $6^2 \times 64$ & s-bottleneck & 6 & 96 & 3 & 1 \\
  $6^2 \times 96$ & s-bottleneck & 6 & 160 & 3 & 2 \\
  $3^2 \times 160$ & s-bottleneck & 6 & 320 & 1 & 1 \\
  $3^2 \times 320$ & s-conv2d & - & 1280 & 1 & 1 \\
  $3^2 \times 1280$ & - & - & - & - & - \\
\hline
\end{tabular}
\caption{
Our stateful MobileNetV2 backbone \cite{Sandler18} for the M=4,N=1 Patchwork configuration. \textbf{s-conv2} is a stateful convolution layer with the Patchwork Cell, as shown in \figref{fig:cell}. \textbf{s-bottleneck} is the stateful MobileNetV2 block with the Patchwork Cell, as shown in \figref{fig:mobilenet}. $t$ is the MobileNetV2 expansion factor as described in \cite{Sandler18}. $c$ is the output channels, $n$ the number of repetition for the layer and $s$ is the stride for the first layer of a kind, repeated layers have the stride 1. The values in this table for the M=2,N=1 and M=4,N=2 setups remain the same, except for doubled height and width in all layers.
}
\label{tbl:mobilenet_architecture}
\end{table}

\section{Patchwork Cell approximation}
\label{sec:Patchwork Cell approximation}


\begin{figure}
  \begin{subfigure}[b]{0.23\textwidth}
    \includegraphics[width=\textwidth]{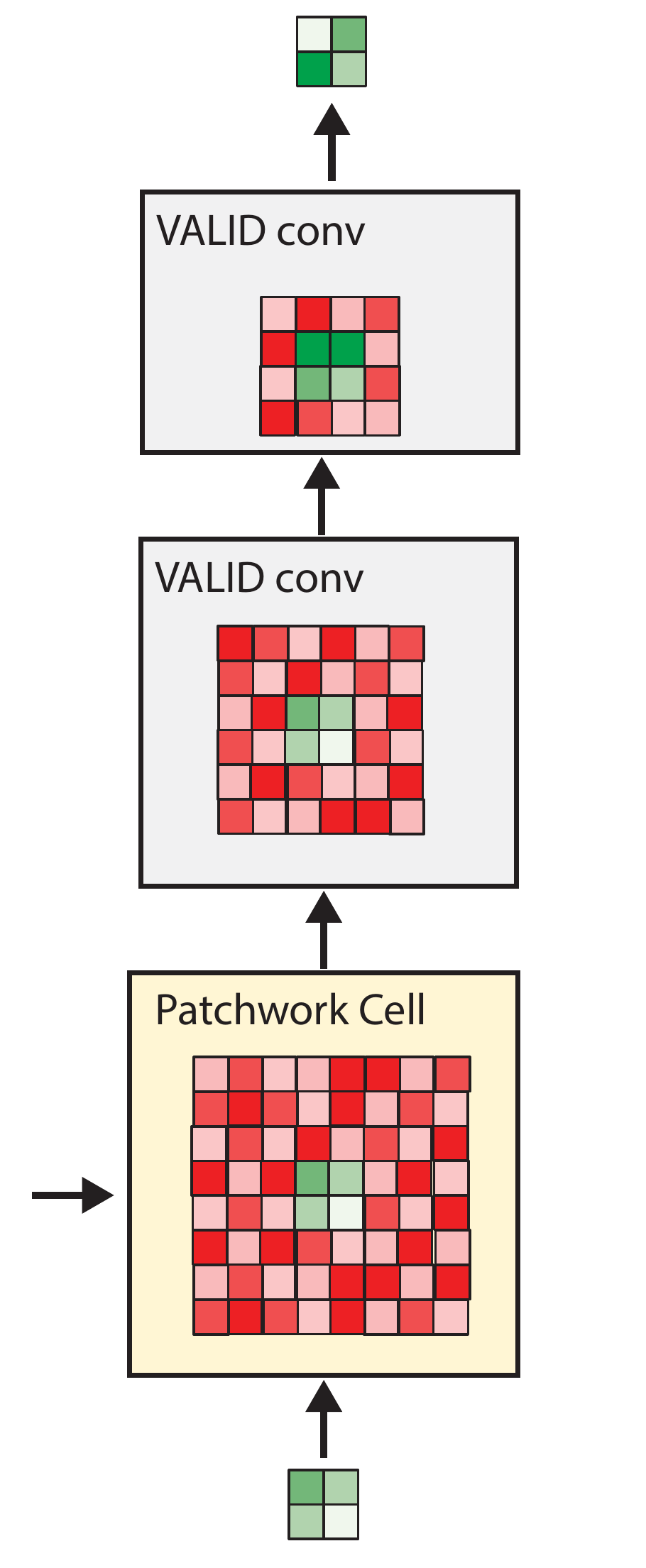}
    \caption{}
    \label{fig:approximation_2}
  \end{subfigure}
  \begin{subfigure}[b]{0.23\textwidth}
    \includegraphics[width=\textwidth]{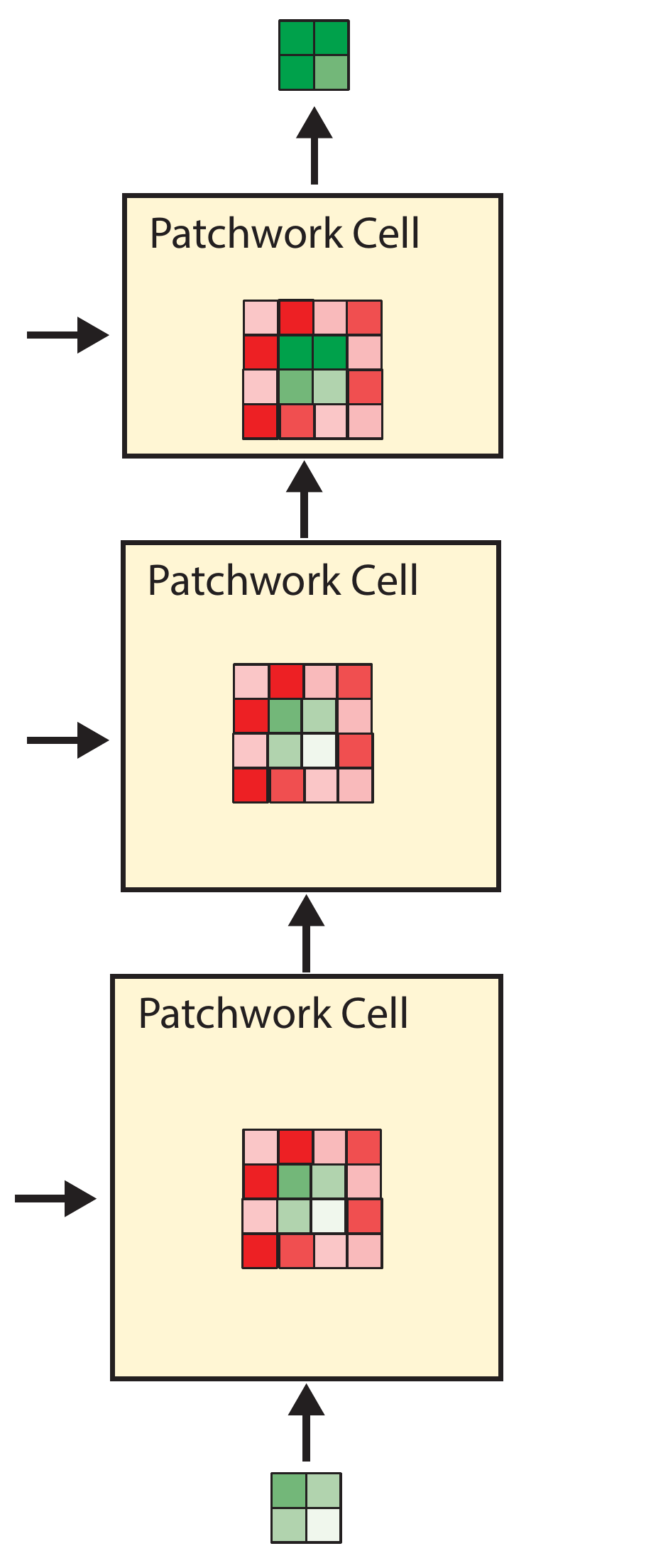}
    \caption{}
    \label{fig:approximation}
  \end{subfigure}
  \caption{
Comparison between different Patchwork Cell usage. Green indicates features extracted from the crop at the current time step. Red indicates features from the past provided by the Patchwork Cell. a) An architecture where the context is approximated by one single Patchwork Cell in the first layer. b): The proposed architecture that gradually approximates the lost context via Patchwork Cells throughout the network.
}
\end{figure}


The Patchwork Cell in \secref{sec:patchwork cell} approximates lost context using features from the past. The most accurate way to do so is to apply a single Patchwork Cell at the start of the network to recover as much context as possible, as shown in \figref{fig:approximation_2}. However, doing this nullifies most of the latency saving that is the motivation of Patchwork. Alternatively, a series of Patchwork Cells can approximate the context incrementally (see \figref{fig:approximation}), which preserves the latency saving while suffers less than 0.1\% of accuracy loss.

\section{Detailed results}
\label{sec:Detailed results}


\tblref{tbl:det_results} and \tblref{tbl:seg_results} show a more detailed view of the quantitative results in \figref{fig:det_results}, \figref{fig:det_results_msecs}, \figref{fig:seg_results}, \figref{fig:seg_results_msecs}. We report the average latency in addition to the maximum latency in both the empirical and theoretical measure. For the empirical latency, we also show the mean and standard deviation. Notably, the standard deviation for any baseline with an interval larger than 1 is high due to the uneven nature of the latency that only occurs every K keyframes. The high variance might be undesirable for many applications. For the maximum empirical latency, we show both the 95 and the 99 percentile. 

\begin{table*}
\begin{center}
\begin{tabular}{|c|l|cc|cc|c|ccc|ccc|ccc|}
\hline
\multirow{2}{*}{ID} & \multirow{2}{*}{Method} & \multicolumn{2}{c|}{MFLOPs}  &
\multicolumn{2}{c|}{msecs}  & \multirow{2}{*}{MAP $\uparrow$} \\ 
\cline{3-6}
                 &                        & max                   & avg & max                   & avg &                       \\ 
\hline
\hline
1 & Single-frame (SF) & 2047  & 2047 & 152.7 / 160.6 & 134.3$\pm$8.9 & 54.7 \\ 
2 & SF interval=4 & 2047 & 512 & 146.9 / 155.0 & 33.8$\pm$58.8 & 53.4 \\ 
3 & SF interval=16 & 2047 & 128 & 127.6 / 147.8 & 8.41$\pm$32.7 & 45.7 \\ 
  4 & Patchwork M=4, N=2, depth=1.4, flip & 1945 & 1945 & 149.2 / 156.3 &
  137.7$\pm$6.2 & \textbf{58.7} \\ 
\hline
5 & Patchwork M=4, N=2, depth=1.4 & 973 & 973 & 75.2 / 79.2 & 68.1$\pm$3.6 & 57.4 \\ 
\hline
6 & SF depth=0.5 & 602 & 602 & 73.4 / 80.9 & 66.3$\pm$3.7 & 47.2 \\ 
7 & SF resolution=0.25 & 512 & 512 & 47.4 / 50.0 & 41.7$\pm$2.6 & 46.8 \\ 
8 & SF interval=4, delay=3 & 512 & 128 & 36.7 / 38.8 & 8.5$\pm$16.9 & 49.3 \\ 
  9 & Patchwork M=4, N=2 & 543 & 543 & 55.2 / 58.4 & 47.9$\pm$3.7 & \textbf{54.3} \\ 
\hline
10 & SF interval=16, delay=15 & 128 & 8 & 8.0 / 9.2 & 0.5$\pm$2.0 & 34.3 \\ 
11 & Patchwork M=4, N=1 & 162 & 162 & 28.1 / 29.8 & 24.2$\pm$2.3 & \textbf{41.6} \\ 
\hline
\end{tabular}
\end{center}
\caption{
  A latency vs. accuracy comparison between the single-frame and
  Patchwork variants ImageNet VID. For the empirical (msecs) metric, we
  report the 95 and 99 percentile latency for \textbf{max} and the mean/std pair for
  \textbf{avg}.
  Methods with similar maximum FLOPs are grouped together.
  Same notation as described in \secref{sec:experiments}.
}
\label{tbl:det_results}
\end{table*}

\begin{table*}
\begin{center}
\begin{tabular}{|c|l|cc|cc|c|c|}
\hline
\multirow{2}{*}{ID} & \multirow{2}{*}{Method} & \multicolumn{2}{c|}{FLOPs}  &
  \multicolumn{2}{c|}{msecs}  & \multirow{2}{*}{$\mathcal{J}$ $\uparrow$} &
  \multirow{2}{*}{$\mathcal{F}$ $\uparrow$}\\ 
\cline{3-6}
  && max & avg & max & avg & & \\ 
\hline
\hline
  1 & Single-frame (SF) & 3307 & 3307 & 190.5 / 205.0 & 174.0$\pm$9.2 & 63.6 & 59.5 \\ 
  2 & SF interval=4 & 3307 & 827 & 171.5 / 181.7 & 41.0$\pm$71.3 & 55.8 & 47.9 \\
  3 & SF interval=16 & 3307 & 207 & 156.5 / 174.3 & 10.3$\pm$40.0 & 40.5 & 31.7 \\
\hline
  4 & SF depth=0.5 & 1240 & 1240 & 108.6 / 114.8 & 99.5$\pm$5.3 & 58.4 & 55.2 \\ 
  5 & SF resolution=0.25 & 827 & 827 & 51.6 / 58.5 & 45.2$\pm$3.5 & 56.6 & 50.8 \\ 
  6& SF interval=4, delay=3 & 827 & 207 & 42.9 / 45.4 & 10.3$\pm$17.8 & 44.6 & 34.1 \\
  7& Patchwork M=4, N=2 & 841 & 841 & 65.5 / 71.1 & 58.5$\pm$3.6 & \textbf{62.1}
  & \textbf{58.8} \\ 
\hline
  8& SF interval=16, delay=15 & 207 & 52 & 9.8 / 10.9 & 0.64$\pm$2.5 & 27.5 & 20.8\\
  9& Patchwork M=4, N=1 & 221 & 221 & 33.4 / 36.2 & 28.3$\pm$2.7 & \textbf{42.2}
  & \textbf{36.9} \\ 
\hline
\end{tabular}
\end{center}
\caption{
Object segmentation results on DAVIS 2016. The experimental setups and markings are similar to those in the detection results, see \tblref{tbl:det_results} for details.}
\label{tbl:seg_results}
\end{table*}

%% file: ms.bbl
\begin{thebibliography}{10}\itemsep=-1pt

\bibitem{Ba15}
J.~Ba, V.~Mnih, and K.~Kavukcuoglu.
\newblock Multiple object recognition with visual attention.
\newblock In {\em ICLR}, 2015.

\bibitem{Bertasius18}
G.~Bertasius, L.~Torresani, and J.~Shi.
\newblock Object detection in video with spatiotemporal sampling networks.
\newblock In {\em ECCV}, 2018.

\bibitem{Caicedo15}
J.~C. Caicedo and S.~Lazebnik.
\newblock Active object localization with deep reinforcement learning.
\newblock In {\em ICCV}, 2015.

\bibitem{Carrasco11}
M.~Carrasco.
\newblock Visual attention: The past 25 years.
\newblock {\em Vision Research}, 2011.

\bibitem{Chen18a}
L.~Chen, A.~Hermans, G.~Papandreou, F.~Schroff, P.~Wang, and H.~Adam.
\newblock Masklab: Instance segmentation by refining object detection with
  semantic and direction features.
\newblock {\em CVPR}, 2018.

\bibitem{Chen18}
L.~Chen, G.~Papandreou, I.~Kokkinos, K.~Murphy, and A.~L. Yuille.
\newblock Deeplab: Semantic image segmentation with deep convolutional nets,
  atrous convolution, and fully connected crfs.
\newblock {\em PAMI}, 2018.

\bibitem{Cheng17}
J.~Cheng, Y.~Tsai, S.~Wang, and M.~Yang.
\newblock Segflow: Joint learning for video object segmentation and optical
  flow.
\newblock In {\em ICCV}, 2017.

\bibitem{Chollet17}
F.~Chollet.
\newblock Xception: Deep learning with depthwise separable convolutions.
\newblock In {\em CVPR}, 2017.

\bibitem{Dai16}
J.~Dai, Y.~Li, K.~He, and J.~Sun.
\newblock {R-FCN:} object detection via region-based fully convolutional
  networks.
\newblock In {\em NIPS}, 2016.

\bibitem{Feichtenhofer17}
C.~Feichtenhofer, A.~Pinz, and A.~Zisserman.
\newblock Detect to track and track to detect.
\newblock {\em ICCV}, 2017.

\bibitem{Girshick15}
R.~Girshick.
\newblock Fast r-cnn.
\newblock In {\em ICCV}, 2015.

\bibitem{Gregor15}
K.~Gregor, I.~Danihelka, A.~Graves, and D.~Wierstra.
\newblock {DRAW:} {A} recurrent neural network for image generation.
\newblock In {\em ICML}, 2015.

\bibitem{Han18}
J.~Han, L.~Yang, D.~Zhang, X.~Chang, and X.~Liang.
\newblock Reinforcement cutting-agent learning for video object segmentation.
\newblock In {\em CVPR}, 2018.

\bibitem{He17}
K.~He, G.~Gkioxari, P.~Doll{\'{a}}r, and R.~B. Girshick.
\newblock Mask {R-CNN}.
\newblock In {\em ICCV}, 2017.

\bibitem{He16}
K.~He, X.~Zhang, S.~Ren, and J.~Sun.
\newblock Deep residual learning for image recognition.
\newblock In {\em CVPR}, 2016.

\bibitem{Ioffe15}
S.~Ioffe and C.~Szegedy.
\newblock Batch normalization: Accelerating deep network training by reducing
  internal covariate shift, 2015.

\bibitem{Jacob18}
B.~Jacob, S.~Kligys, B.~Chen, M.~Zhu, M.~Tang, A.~G. Howard, H.~Adam, and
  D.~Kalenichenko.
\newblock Quantization and training of neural networks for efficient
  integer-arithmetic-only inference.
\newblock {\em CVPR}, 2018.

\bibitem{Jain17}
S.~D. Jain, B.~Xiong, and K.~Grauman.
\newblock Fusionseg: Learning to combine motion and appearance for fully
  automatic segmentation of generic objects in videos.
\newblock In {\em CVPR}, 2017.

\bibitem{Jayaraman18}
D.~Jayaraman and K.~Grauman.
\newblock End-to-end policy learning for active visual categorization.
\newblock {\em PAMI}, 2018.

\bibitem{Koh17}
Y.~J. Koh and C.~Kim.
\newblock Primary object segmentation in videos based on region augmentation
  and reduction.
\newblock In {\em CVPR}, 2017.

\bibitem{Krizhevsky12}
A.~Krizhevsky, I.~Sutskever, and G.~E. Hinton.
\newblock Imagenet classification with deep convolutional neural networks.
\newblock In {\em NIPS}, 2012.

\bibitem{Lebedev14}
V.~Lebedev, Y.~Ganin, M.~Rakhuba, I.~Oseledets, and V.~Lempitsky.
\newblock Speeding-up convolutional neural networks using fine-tuned
  cp-decomposition.
\newblock \emph{arXiv preprint arXiv:1412.6553}, 2014.

\bibitem{Li18}
Y.~Li, J.~Shi, and D.~Lin.
\newblock Low-latency video semantic segmentation.
\newblock In {\em CVPR}, 2018.

\bibitem{Liu18}
M.~Liu and M.~Zhu.
\newblock Mobile video object detection with temporally-aware feature maps.
\newblock {\em CVPR}, 2018.

\bibitem{Liu16}
W.~Liu, D.~Anguelov, D.~Erhan, C.~Szegedy, S.~E. Reed, C.~Fu, and A.~C. Berg.
\newblock {SSD:} single shot multibox detector.
\newblock In {\em ECCV}, 2016.

\bibitem{Markant14}
J.~Markant and D.~Amso.
\newblock Leveling the playing field: Attention mitigates the effects of
  intelligence on memory.
\newblock {\em Cognition}, 2014.

\bibitem{Mnih14}
V.~Mnih, N.~Heess, A.~Graves, and K.~Kavukcuoglu.
\newblock Recurrent models of visual attention.
\newblock In {\em NIPS}, 2014.

\bibitem{Mnih15}
V.~Mnih, K.~Kavukcuoglu, D.~Silver, A.~A. Rusu, J.~Veness, M.~G. Bellemare,
  A.~Graves, M.~Riedmiller, A.~K. Fidjeland, G.~Ostrovski, S.~Petersen,
  C.~Beattie, A.~Sadik, I.~Antonoglou, H.~King, D.~Kumaran, D.~Wierstra,
  S.~Legg, and D.~Hassabis.
\newblock Human-level control through deep reinforcement learning.
\newblock {\em Nature}, 2015.

\bibitem{Perazzi16}
F.~Perazzi, J.~Pont-Tuset, B.~McWilliams, L.~Van~Gool, M.~Gross, and
  A.~Sorkine-Hornung.
\newblock A benchmark dataset and evaluation methodology for video object
  segmentation.
\newblock In {\em CVPR}, 2016.

\bibitem{Redmon15}
J.~Redmon, S.~K. Divvala, R.~B. Girshick, and A.~Farhadi.
\newblock You only look once: Unified, real-time object detection.
\newblock {\em CVPR}, 2016.

\bibitem{Ren17}
S.~Ren, K.~He, R.~B. Girshick, and J.~Sun.
\newblock Faster {R-CNN:} towards real-time object detection with region
  proposal networks.
\newblock {\em PAMI}, 2017.

\bibitem{Russakovsky15}
O.~Russakovsky, J.~Deng, H.~Su, J.~Krause, S.~Satheesh, S.~Ma, Z.~Huang,
  A.~Karpathy, A.~Khosla, M.~Bernstein, A.~C. Berg, and L.~Fei{-}Fei.
\newblock Imagenet large scale visual recognition challenge.
\newblock {\em IJCV}, 2015.

\bibitem{Sandler18}
M.~Sandler, A.~G. Howard, M.~Zhu, A.~Zhmoginov, and L.~Chen.
\newblock Inverted residuals and linear bottlenecks: Mobile networks for
  classification, detection and segmentation.
\newblock In {\em CVPR}, 2018.

\bibitem{Shelhamer16}
E.~Shelhamer, K.~Rakelly, J.~Hoffman, and T.~Darrell.
\newblock Clockwork convnets for video semantic segmentation.
\newblock In {\em ECCV}, 2016.

\bibitem{Simonyan15}
K.~Simonyan and A.~Zisserman.
\newblock Very deep convolutional networks for large-scale image recognition.
\newblock In {\em ICLR}, 2015.

\bibitem{Song18}
H.~Song, W.~Wang, S.~Zhao, J.~Shen, and K.-M. Lam.
\newblock Pyramid dilated deeper convlstm for video salient object detection.
\newblock In {\em ECCV}, 2018.

\bibitem{Sutton98}
R.~S. Sutton and A.~G. Barto.
\newblock {\em Introduction to Reinforcement Learning}.
\newblock MIT Press, 1998.

\bibitem{Szegedy16}
C.~Szegedy, V.~Vanhoucke, S.~Ioffe, J.~Shlens, and Z.~Wojna.
\newblock Rethinking the inception architecture for computer vision.
\newblock In {\em CVPR}, 2016.

\bibitem{Tokmakov17a}
P.~Tokmakov, K.~Alahari, and C.~Schmid.
\newblock Learning motion patterns in videos.
\newblock In {\em CVPR}, 2017.

\bibitem{Tokmakov17}
P.~Tokmakov, K.~Alahari, and C.~Schmid.
\newblock Learning video object segmentation with visual memory.
\newblock In {\em ICCV}, 2017.

\bibitem{vanHasselt16}
H.~van Hasselt, A.~Guez, and D.~Silver.
\newblock Deep reinforcement learning with double q-learning.
\newblock {\em AAAI}, 2016.

\bibitem{Xiao18}
F.~Xiao and Y.~J. Lee.
\newblock Video object detection with an aligned spatial-temporal memory.
\newblock In {\em ECCV}, 2018.

\bibitem{Zhou17}
A.~Zhou, A.~Yao, Y.~Guo, L.~Xu, and Y.~Chen.
\newblock Incremental network quantization: Towards lossless cnns with
  low-precision weights.
\newblock In {\em ICLR}, 2017.

\bibitem{Zhu18}
X.~Zhu, J.~Dai, L.~Yuan, and Y.~Wei.
\newblock Towards high performance video object detection.
\newblock In {\em CVPR}, 2018.

\bibitem{Zhu18a}
X.~Zhu, J.~Dai, X.~Zhu, Y.~Wei, and L.~Yuan.
\newblock Towards high performance video object detection for mobiles.
\newblock \emph{arXiv preprint arXiv:1804.05830}, 2018.

\end{thebibliography}
